\documentclass[runningheads]{llncs}

\usepackage{eccv}

\usepackage{eccvabbrv}

\usepackage{graphicx}
\usepackage{booktabs}

\usepackage[accsupp]{axessibility}  
\usepackage{algorithm}
\usepackage{algpseudocode}

\usepackage{dsfont}

\usepackage{hyperref}

\usepackage{orcidlink}

\begin{document}

\title{Discovering Geometric Biases in 3D Face Reconstruction: A Curvature-Aware Spectral Framework for Fairness Evaluation} 


\titlerunning{Discovering Geometric Biases in 3D Face Reconstruction}

\author{Veronika Shilova\inst{1,2} \and
Emmanuel Malherbe\inst{1} \and
Giovanni Palma\inst{3} \and Panagiotis-Alexandros Bokaris\inst{3} \and Laurent Risser\inst{2} \and Jean-Michel Loubes\inst{2,4,5}}

\authorrunning{V.~Shilova et al.}

\institute{Artefact Research Center, Paris, France \and Institut de Mathématique de Toulouse, CNRS, Toulouse, France \and
L'Oréal Recherche et Innovation, Paris, France
 \and
Université de Toulouse, ANITI, Toulouse, France \and
INRIA Regalia, Bordeaux, France}

\maketitle

\begin{abstract}

3D Morphable Models (3DMMs) remain the standard parametric shape priors for many state-of-the-art 3D face reconstruction algorithms. However, as these models are derived from a finite number of 3D face samples, they inherit the morphological biases of their training data, potentially limiting their generalizability across diverse global populations. 
In this paper, we propose a novel framework to analyze 3DMM reconstructions through the lens of surface curvature, with the objective to discover, quantify and visualize biases.
While standard evaluation metrics often rely on Euclidean distances, our  reconstruction error captures subtle surface nuances such as local topology or undulations.
To do so, we leverage the Laplace-Beltrami Operator (LBO) to generate high-resolution curvature error maps, providing a localized and geometrically meaningful visualization of discrepancies between ground truth faces and reconstructed meshes. 
We derive from it an error metric that we validated through a user study, observing a significantly higher correlation to human perception compared to traditional methods.
Furthermore, we conduct extensive experiments across several 3DMM bases and fitting algorithms, uncovering systematic age-related biases and providing preliminary evidence of biases associated with gender and ethnicity. Our findings highlight the necessity of adopting curvature-aware evaluation protocols to ensure demographic fairness and geometric precision in future 3D face reconstruction research. The code and annotation data are available at \url{https://github.com/artefactory/3dface-fairness}.



\end{abstract}

\section{Introduction}
\label{sec:intro}


The field of 3D face reconstruction from images has become important in computer vision, as it makes it possible to compute accurate representations of the human facial morphology, with many applications such as in the field of cinema \cite{gerogiannis2025arc2avatar}, virtual reality \cite{lombardi2018deep,thies2016facevr}, cosmetics \cite{yang2024makeup}, biometric security \cite{leelavathi2025enhanced} or medicine \cite{mueller2011missing}. 
An underlying component of any face reconstruction is the parametrization of the 3D shapes, for which 3D Morphable Models (3DMM) \cite{10.1145/311535.311556} have established themselves as fundamental parametric shape priors for 3D face reconstruction \cite{egger20203d}. 

\begin{figure}[tb]
    \centering
    \includegraphics[width=\linewidth]{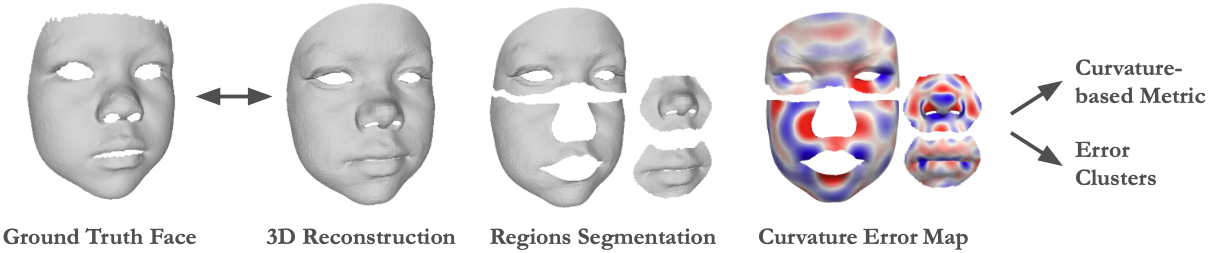}
    \caption{Overview of our curvature-aware evaluation framework for bias discovery. Given a ground truth scan and its corresponding 3D reconstruction, we perform an anatomical segmentation of the reconstructed mesh. We compute Curvature-Based Error Maps using the Laplace-Beltrami Operator to isolate localized geometric discrepancies that Euclidean metrics often overlook. By performing spectral clustering on these error maps, we identify distinct failure modes and correlate them with demographic factors.} 
    \label{fig:overview-figure}
\end{figure}


The 3DMM formalism is based on two fundamental assumptions. 
First, it establishes a dense, point-to-point correspondence across registered face scans in its database. This shared topology allows for the generation of anatomically plausible identities through linear combinations of basis shapes. Second, the model explicitly disentangles facial shape and texture, and separates them from extrinsic scene parameters, such as illumination and camera parameters.
The classic 3DMM formulation decomposes facial geometry and texture using Principal Component Analysis (PCA) bases derived on a finite set of 3D face scans. 

In this sense, they are prone to machine-learning bias issues \cite{Besse03042022,shilova2025fairness,cui2024boosting}, meaning that they can systematically offer poor performances on specific categories of faces, due to, \eg, age, gender, or geographical origin. 
These limitations can have a negative impact on the reputation of companies using these tools and can even be considered as  discriminatory in certain cases. Strategies for detecting and quantifying these biases are therefore necessary to ensure full confidence in these tools. 
We want to emphasize that regulatory texts such as the \textit{AI act} \cite{AIact2024}, explicitly mention the need for accuracy metrics, as \eg, in its Article 15: \textit{The levels of accuracy and the relevant accuracy metrics of high-risk AI systems shall be declared in the accompanying instructions of use.} To the best of our knowledge, this work represents the first comprehensive evaluation of demographic bias in 3D face reconstruction, addressing a critical gap in a field where facial analysis is increasingly categorized as high-risk.

In our paper, we explore new solutions to quantify and detect algorithmic biases in 3D face reconstruction algorithms (\cref{fig:overview-figure}). Our main contributions are:
\begin{itemize}
    \item \textit{Curvature-Aware Evaluation Framework:} We introduce a new framework for 3D face reconstruction error analysis that focuses on local differential geometry using the Laplace-Beltrami Operator (LBO), to quantify and visualize surface discrepancies that are often invisible to standard Euclidean metrics.
    \item \textit{Perceptual Validation:} We demonstrate through a user study that our curvature-based reconstruction error aligns more closely with human qualitative judgment than traditional point-wise distances.
    \item \textit{Systematic Bias Discovery:} To our knowledge, we conduct the first study to expose morphological biases in 3D face reconstruction algorithms. Our analysis reveals substantial fidelity gaps across demographic factors, specifically identifying systematic age-related biases and providing preliminary evidence of gender- and ethnicity-related biases.
\end{itemize}


While demonstrated on 3DMMs, our curvature-aware evaluation framework is agnostic to the reconstruction method, and our bias discovery strategy generalizes to any template-based 3D surface reconstruction.

\section{Related Work}
\label{sec:related_work}

\subsection{3D Morphable Model (3DMM) and 3D Face Reconstruction}

The objective of 3D face reconstruction is to recover the dense geometry of a face from a single RGB(-D) image or an image collection of the same person. To do so, most methods use a 3DMM as its shape prior. Consequently, the reconstruction task is formulated as a regression or optimization problem to estimate the latent shape parameters. Current approaches either use deep networks to predict them \cite{deng2019accurate, zielonka2022towards, wang20243d, feng2021learning, shang2020self, sanyal2019learning} or optimize a combination of energy terms \cite{blanz2003face, hu2017avatar, thies2016face2face, yamaguchi2018high}, such as landmark-based alignment and photometric consistency, often subject to various statistical regularizations.

Basel Face Model (BFM) \cite{paysan20093d} was the first publicly available 3DMM trained on 200 individuals. More recently, FLAME \cite{li2017learning} has become the standard for its database diversity (3800 face scans) and its ability to model the head, neck, and jaw articulation more realistically than previous vertex-based models. However, since these models were trained on limited amount of data, they often fail to capture the fine-grained geometric details that differentiate individuals across diverse demographics.


\subsection{3D Face Reconstruction Metrics}
Standard evaluation in the field typically relies on global geometric fidelity. The most prevalent metric is point-to-point or point-to-surface Root Mean Square Error (RMSE) or Normalized Mean Square Error (NMSE) \cite{feng2018evaluation}, often calculated using the Iterative Closest Point (ICP) \cite{besl1992method,amberg2007optimal} algorithm for alignment. While these metrics assess the global "shell" of the face, they are insensitive to regional anatomical properties.

To try to bridge the gap, \cite{chai2022realy} proposed a new benchmark called REALY consisting of 100 globally aligned high-quality face scans with 4 anatomical region masks (including nose, mouth, cheek and forehead) and a bi-directional evaluation pipeline with Non-Rigid Iterative Closest Point \cite{besl1992method,amberg2007optimal} for better alignment. Moreover, \cite{sariyanidi20253d} introduced Modularized 3D Face reconstruction Benchmark (M3DFB) where they separate the fundamental components of error computation such as rigid alignment or point correspondence, allowing
one to quantify the effect of each. Nevertheless, they still focus only on Euclidean distances between meshes in their evaluation pipelines.

While bias in facial analysis is well-documented in 2D recognition \cite{grother2019face,cook2019demographic}, its quantification in 3D reconstruction has not been widely studied in the literature. Recent attempts to quantify bias \cite{cui2024boosting} were limited by reliance on global Euclidean distances, resulting in marginal and ultimately inconclusive performance gaps across demographics. This highlights the need for curvature-aware metrics capable of uncovering localized reconstruction failures.

\section{Methodology}
\subsection{Background and Notations}
\subsubsection{Face Mesh.}
Let $\mathcal{M} = \{\mathcal{V}, \mathcal{F}\}$ be a triangle mesh with vertex positions $\mathcal{V}$ in the 3D ambient space,   and a list of triangular faces $\mathcal{F}$. 
In this paper, $\mathcal{M}$ generally represents the surface of someone's face, however we can equivalently consider a subpart of the mesh $R_{\mathcal{M}} \subsetneq \mathcal{M}$ constituting a meaningful morphological region of the face (\eg, nose), also called region-of-interest.
Let $\mathcal{M}'$ be another mesh that aims at representing the same face, possibly with a different number of vertices $|\mathcal{V}'|\neq |\mathcal{V}|$. 
In the context of our paper, dedicated to 3D surface reconstruction evaluation, the ground-truth mesh is $\mathcal{M}$ and its reconstruction is $\mathcal{M}'$.
In all the following, we assume that meshes $\mathcal{M}$ and $\mathcal{M}'$ are first rigidly aligned using Iterative Closest Point algorithm \cite{besl1992method}.
%
\subsubsection{Mesh Mapping.}
We denote by $T_{\mathcal{M}' \rightarrow \mathcal{M}}$ a mapping between $\mathcal{M}'$ and $\mathcal{M}$, where the subscript represents the direction of the map and $T_{\mathcal{M}' \rightarrow \mathcal{M}}(v)$ gives the coordinates of the mapped position of a vertex $v \in \mathcal{M}'$ on shape $\mathcal{M}$. We consider two different types of maps: 
the point-to-surface map $T^{pts}_{\mathcal{M}' \rightarrow \mathcal{M}}$ (also called vertex-to-plane or vertex-to-point), which is the most commonly used, and the vertex-to-vertex map $T^{vtx}_{\mathcal{M}' \rightarrow \mathcal{M}}$, which is less accurate but remains considered in some applications. Specifically, $T^{vtx}_{\mathcal{M}' \rightarrow \mathcal{M}}$ maps each vertex in shape $\mathcal{M}'$ to the closest vertex in $\mathcal{V}$, and $T^{pts}_{\mathcal{M}' \rightarrow \mathcal{M}}$ maps each vertex in shape $\mathcal{M}'$ to the closest point in a face of $\mathcal{F}$. We will only use the superscript to disambiguate the two maps when necessary.




\subsubsection{Normalized Mean Square Error (NMSE).} The NMSE metric, denoted as $\mathcal{E}^{pts}_{NMSE}$, computes the distance between two meshes $\mathcal{M}$ and $\mathcal{M}'$ using the vertex-to-point  map $T^{pts}_{\mathcal{M} \rightarrow \mathcal{M}'}$:

\begin{equation}
\label{eq:nmse-vtx}
    \mathcal{E}^{pts}_{NMSE}(\mathcal{M},\mathcal{M}') = \frac{1}{|\mathcal{V}'|} \sum_{v \in \mathcal{M}'}|| v - T^{pts}_{\mathcal{M}' \rightarrow \mathcal{M}}(v) ||^2,
\end{equation}
where $|\mathcal{V}'|$ is the number of vertices in shape $\mathcal{M}'$. We similarly define $\mathcal{E}^{vtx}_{NMSE}$, substituting point-to-surface mapping  for the vertex-to-vertex correspondence.
We recall that such similarity metrics are routinely used to assess the accuracy of 3D surface reconstructions 
\cite{besl1992method,amberg2007optimal,feng2018evaluation}.

\subsubsection{3DMM.} 

A face $\mathcal{M} = (\mathcal{V}, \mathcal{F})$ is expressed in the 3DMM base if it verifies $\mathcal{V} = \vec{\mu} + \sum_{j=1}^{n} \beta_j \vec{\psi_j}$, with $\vec{\mu}$ the mean shape, $\vec \psi_j$ the principal components and ${\beta_j}$ the shape parameters of the 3DMM. 
In this paper, we assume that $\mathcal{M}'$ is expressed in a given 3DMM base. In such base, all vertices systematically represent the same points semantically. Note that all our methods would equivalently apply if we use a standard base that do not originates from a 3DMM, or if it is the mesh $\mathcal{M}$ that is expressed in the 3DMM.


\subsection{Quantifying and Visualizing the Geometric Reconstruction Accuracy Using Curvature Reconstruction Errors (CRE)}

To quantify the preservation of structural identity and identify systematic geometric biases, we propose to use the \textbf{Mean Curvature} ($H$), a second-order differential property characterizing the local geometry of the manifold. Unlike point-wise Euclidean metrics, $H$ is sensitive to the local surface ``character''---such as ridges, valleys, and high-frequency morphological details---making it a robust indicator of reconstruction fidelity.

\subsubsection{Mathematical Definition of the Mean Curvature.}
For a smooth surface $\mathcal{S} \subset \mathbb{R}^3$, the mean curvature $H$ at a surface point $(x,y,z)$ is defined as the arithmetic mean of the two principal curvatures, $\kappa_1$ and $\kappa_2$:
\begin{equation}
H = \frac{\kappa_1 + \kappa_2}{2} \,.
\end{equation}
In the context of differential geometry, $H$ can also be derived using the Laplace-Beltrami operator, which is the manifold generalization of the Laplacian to curved surfaces \cite{botsch2010polygon}. By writing this operator $\Delta$ and the parametrized coordinate function $\vec{p}: (\xi, \eta) \in \mathcal{S} \rightarrow \big(x\left(\xi,\eta\right), y\left(\xi,\eta\right), z\left(\xi,\eta\right)\big) \in \mathbb{R}^3$ defined on surface $\mathcal{S}$, we have
\begin{equation}
    H  = -\frac{1}{2}\langle \vec{n}, \Delta \vec{p} \rangle \,,
    \label{eq:mean-curvature-laplace}
\end{equation}
where $\vec n$ is the unit normal vector at the considered point. In the following, we will describe a numerical estimation of $H$ following this equation.


\subsubsection{Laplace-Beltrami Estimation using Local Quadratic Regressions.}
Since triangular meshes are piecewise linear, they lack well-defined analytic derivatives at the vertices. 
Existing methodologies to estimate the curvature \cite{hamann1992curvature, meek2000surface, gatzke2003assessing} starts by fitting a quadratic bivariate surface to each vertex $\vec{v}_i \in \mathcal{V}$ and its neighbor vertices.
We propose to differ by locally estimating the Laplace-Beltrami operator, in order to capture the underlying differential properties and apply the Equation \ref{eq:mean-curvature-laplace}. Indeed, such operator will be independent from the distances to tangent plane, and thus less prone to noise in our data, such as the one related to the tangent plane and its normal. In addition, we propose to use a numerical estimation of this estimator, performed on local neighborhoods that are adapted to user-defined parameters.


For each vertex $\vec{v}_i$, we consider its $R$-ring neighborhood $\mathcal{N}_R(\vec{v}_i) = \{ \vec{v} \in \mathcal{V} \mid d_G(\vec{v}_i, \vec{v}) \leq R \}$, where $d_G(\vec{v}_i, \vec{v})$ denotes the graph distance (the shortest path in terms of number of edges) between $\vec{v}_i$ and $\vec{v}$. We compute the corresponding surface normal $\vec n_i$ and the tangent plane, written in coordinates $(\xi, \eta)$. We consider a scalar function defined on this plane $f: (\xi, \eta) \in \mathbb{R}^2 \rightarrow \mathbb{R}$. For instance, the coordinate function $\vec p$ (Equation \ref{eq:mean-curvature-laplace}) is locally a concatenation of $3$ such functions.
We write $\hat f_{\vec a}$ the second-order polynomial approximation of $f$:
\begin{equation}
\hat f_{\vec{a}}(\xi, \eta) = a_1 \xi^2 + a_2 \xi\eta + a_3 \eta^2 + a_4 \xi + a_5 \eta + a_6 \,.
\end{equation}
where the vector of coefficients $\vec{a}=[a_1, a_2, a_3, a_4, a_5, a_6]^\top \in \mathbb{R}^6$ are obtained by solving the least squares problem on the local neighborhood of $v_i$:
\begin{equation}
\min_{\vec{a}} \sum_{v_j \in \mathcal{N}_R(v_i)} \left( \hat f_{\vec{a}}\left(\xi_{v_j}, \eta_{v_j}\right) - f_{v_j}\right)^2 \,,
\end{equation}
where we write $(\xi_{v_j}, \eta_{v_j})$ the projected coordinates of the neighbor $v_j$ in the tangent plane, and $f_{v_j}=f(\xi_{v_j}, \eta_{v_j})$.
This minimization can be rewritten in the form $\min_{\vec{a}} \|\mathbf{X}\vec{a} - \vec{f}\|^2$. It has an analytical solution $\vec{a} = \mathbf{C}(v_i)\vec{f}$, where $\mathbf{C}(v_i) = (\mathbf{X}^\top \mathbf{X})^{-1} \mathbf{X}^\top$ is calculated from the neighbors $v_j$ of vertex $v_i$. This matrix $\mathbf{C}(v_i)$ acts as the pseudoinverse that maps a function's values $f_{v_j}$ directly to the  coefficients $a_m$ of its polynomial approximation $f_{\vec a}$, and can be used to approximate the Laplace operator at the origin $(0, 0)$:
%
%
%
\begin{equation}
  \Delta \vec{f} = \frac{\partial^2 f}{\partial \xi^2} + \frac{\partial^2 f}{\partial \eta^2}
  \approx 2a_1 + 2a_3 = \sum_{v_j \in \mathcal{N}_R(v_i)} 2(C_{1j} + C_{3j}) f_{v_j}.
    \label{eq:laplace-regression-fit}
\end{equation}
Thus, we can approximate Laplace operator $\Delta$ with the sparse matrix $\tilde{\Delta} \in \mathbb{R}^{|\mathcal{V}| \times |\mathcal{V}|}$ with entries $\tilde{\Delta}_{ij}= 2 (C(v_i)_{1j} + C(v_i)_{3j})$ when $i,j$ verifies $v_j \in \mathcal{N}_R(v_i)$ ($\tilde{\Delta}_{ij}=0$ otherwise). Following the fundamental identity in \cref{eq:mean-curvature-laplace} from differential geometry, we calculate the mean curvature at each vertex $v_i$ as follows
%
%
%
%
\begin{equation}
    H^\mathcal{M}(v_i) = -\frac{1}{2} \langle \vec n_i ,  ( \tilde \Delta v )_i \rangle \,,
\end{equation}
where $\vec n_i \in \mathbb{R}^3$ is the unit normal vector at the vertex ${v}_i$, 
$v \in \mathbb{R}^{|\mathcal{V}| \times 3}$ is the matrix of the Cartesian coordinates of the vertices $\mathcal{V}$, and $( \tilde \Delta v )_i \in \mathbb{R}^3$ is the local Laplace-Beltrami estimation of each $\vec p$ dimension.

Please note that the sparse matrix $\tilde \Delta$ is nonsymetric and can be efficiently computed since the matrices $\mathbf{C}(v_i)$ are of small size. 
The user-defined parameter $R$ provides a trade-off between local precision and numerical stability when estimating the curvature.
While a larger $R$-ring neighborhood enhances the robustness of the operator against discretization noise, it introduces a smoothing effect that may attenuate high-frequency geometric details.

\subsubsection{Curvature-Based Error Metric.}
Following this formulation, we introduce the \textbf{Curvature Reconstruction Error (CRE)}. Given a ground-truth mesh $\mathcal{M}$ and its reconstruction $\mathcal{M}'$, we define the curvature-based error map $\epsilon_H (v_i) \in \mathbb{R}^{|\mathcal{V}'|}$ as:
%
%
\begin{align}
    \label{eq:error-map}
    \epsilon_H ({v}_i) =  H^{\mathcal{M}'}\big({v}_i\big) - H^{\mathcal{M}}\big(T^{pts}_{ \mathcal{M}'\rightarrow{}\mathcal{M}}({v}_i)\big) \,,
\end{align}
where $H^{\mathcal{M}}(.)$ is computed on the vertices of $\mathcal{M}$ using the methodology described in the previous section, and linearly interpolated on the triangular faces $\mathcal{F}$ for the positions given by the mapping $T^{pts}_{ \mathcal{M}'\rightarrow{}\mathcal{M}}$.

We derive from $\epsilon_H$ a global error metric as a scalar quantity
\begin{align}
    \label{eq:cre-metric}
    \mathcal{E}_{CRE}(\mathcal{M},\mathcal{M}') = & \frac{1}{Tr(M)} \|M\epsilon_H\|_{L_1} \,,
\end{align}
where $M$ is the diagonal mass matrix whose entries $M_{i,i}$ correspond to the barycentric area for each vertex $v_i \in \mathcal{V}'$, $\|\cdot\|_{L_1}$ denotes the $L_1$ norm and $Tr(\cdot)$ the trace.
%

\subsection{Unknown Geometric Bias Detection Using the Laplace-Beltrami Operator}
In this section, we leverage the spectral decomposition induced from Laplace-Beltrami operator and use it to identify various failure modes of 3D face reconstruction methods.

\subsubsection{Spectral Decomposition.} Manifold harmonics (also called shape harmonics) are defined as the eigenfunctions $\phi_k$ of the Laplace-Beltrami operator $\Delta$ introduced in the previous section: 
\begin{equation}
    \Delta \phi_k = \lambda_k \phi_k, \, \, k=1, \ldots , |\mathcal{V}|.
\end{equation}

These eigenfunctions form a spectral basis tailored to the specific geometry of the manifold, where lower eigenvalues $\lambda_k$ correspond to smooth, global deformations while higher ones correspond to rapidly oscillating, local details. Please note that this spectral decomposition would be difficult to be computed using the estimated $\tilde \Delta$ matrix introduced in the previous section, since it is not a symmetric matrix. Moreover, the Laplace-Beltrami operator was previously estimated mesh-wise, while the operator below will be the same across meshes of a 3DMM.

For a discrete triangular mesh $\mathcal{M} = (\mathcal{V}, \mathcal{F})$, the eigenvectors and eigenvalues are obtained by solving the generalized eigenvalue problem
\begin{equation}
\label{eq:lbo-generalized-eigenproblem}
    L \vec{\phi_k} = \lambda_k M \vec{\phi_k},
\end{equation}
where $L$ is the cotangent weight (stiffness) matrix representing the weak Laplacian, and $M$ is the diagonal mass matrix defined above. $M^{-1}L$ corresponds to the analytical expression of the Laplace-Beltrami operator for 1-ring neighborhoods, and is symmetrical. The resulting eigenvectors $\vec{\phi_k} \in \mathbb{R}^{|\mathcal{V}|}$ are orthonormal with respect to the mass matrix, meaning $\vec{\phi_k}^\top M \vec{\phi_{k'}} = \delta_{kk'}$. 
These eigenvectors can be easily precomputed for a mesh $\mathcal{M}$ using standard numerical packages \cite{lehoucq1998arpack, sharp2020laplacian}.

\subsubsection{Spectral Error Projection for a 3DMM.} 
We propose to leverage the shape harmonics in order to re-express the reconstruction error into geometric frequencies, instead of scalars defined at the vertices level. To do so, we solve the eigenvalues problem Eq.~\eqref{eq:lbo-generalized-eigenproblem} for the mesh $\mathcal{M}$ -- or region mesh $R_{\mathcal{M}}$ -- from the 3DMM template $\vec \mu$, and its corresponding M and L matrices. Please note that, contrary to curvature-based error computation that is mesh-wise, we now use a general Laplace-Beltrami operator, whose the corresponding spectral projection applies to any face expressed in the 3DMM.

Given a user-defined parameter $K$, let $\Phi := [\vec{\phi_k}] \in \mathbb{R}^{|\mathcal{V}| \times K}$ be the orthonormal basis with respect to the mass matrix M, with eigenvalues in increasing order $0 \leq \lambda_1 < \dots < \lambda_K$. Given a curvature-based reconstruction error $\epsilon_H \in \mathbb{R}^{|\mathcal{V}|}$, we can compute its spectral coefficients $\vec{c}_{\epsilon_H} \in \mathbb{R}^K$ via the spectral projection
\begin{equation}
\label{eq:spectral-proj-lbo}
    \vec{c}_{\epsilon_H} = \Phi^\top M \epsilon_H.
\end{equation}
%





\subsubsection{Unknown Bias Discovery using Clustering  in the Spectral Space.}
\label{sec:k-means-lbo}

\begin{algorithm}[t]
\caption{Clustering of Error Maps}
\label{alg:kmeans-algo}
    \begin{algorithmic}[1]
    \Require Template mesh $\vec{\mu}$, set of error maps $\{\epsilon^{(n)}_H\}_{n=1}^N$, maximum frequency $K\in \mathbb{N}$
    \State Calculate cotangent matrix $L$ from template mesh $\vec{\mu}$
    \State Calculate mass matrix $M$ from template mesh $\vec{\mu}$
    \State Find $\Phi = [\vec{\phi_k}]$, $\lambda_k$, $k=1, \dots, K$, from \cref{eq:lbo-generalized-eigenproblem}
    \For{$n=1$ to $N$}
        \State Project error map to spectral space of template: $\vec{c^{(n)}_{\epsilon_H}} \gets$ $\Phi^\top M \epsilon^{(n)}_H$
    \EndFor
    \State Perform K-means on the set $\{\vec{c^{(n)}_{\epsilon_H}}\}_{n=1}^N$
    \For{each centroid $\vec{\bar{c}}_{\epsilon_H}$}
        \State Project back to vertex space: $\bar{\epsilon}_H \gets \Phi \vec{\bar{c}}_{\epsilon_H}$
    \EndFor
    \State \Return $\bar{\epsilon}_H$, centroids in the vertex space
\end{algorithmic}
\end{algorithm}

We now consider a set of $N$ pairs of ground truth faces with their reconstruction $\{(\mathcal{M}^{(n)}, \mathcal{M}'^{(n)})\}_{n=1}^N$, assuming that they sufficiently represent the possible biases of the underlying reconstruction algorithm.
We propose a clustering method that aims at discovering systematic failure modes.
To do so, this method identifies recurring patterns of geometric bias by analyzing the curvature-based reconstruction error maps $\epsilon_H$ of these $N$ pairs.

With this objective, a practical challenge is that the number of ground truth faces $N$ is generally much lower than the dimension of $\epsilon_H$, which is $|\mathcal{V}'|$. As described in Algorithm \ref{alg:kmeans-algo}, we project the error maps into the spectral space (Equation \ref{eq:spectral-proj-lbo}), and perform a
K-means clustering \cite{macqueen1967some} on the spectral errors $\{\mathbf{c}_{\mathcal{\epsilon_H}}^{(n)}\}_{n=1}^N$. Indeed, this spectral space is of much lower dimension $K \ll |\mathcal{V}'|$ and still well represents the geometric patterns of the errors. 

Given a resulting cluster, its centroid can be easily visualized by reconstructing the signal in the spatial domain through the inverse transform:
\begin{equation}
\label{eq:reconstruction}
    \hat{\epsilon}_H = \Phi \vec{c}_{\epsilon_H}.
\end{equation}
which can be used for visualization purposes to illustrate the failure mode of the cluster. 



\section{Experiments}
\subsection{Protocol}
\subsubsection{Ground Truth Faces.}
\begin{figure}[htb]
  \centering
    \includegraphics[width=\textwidth]{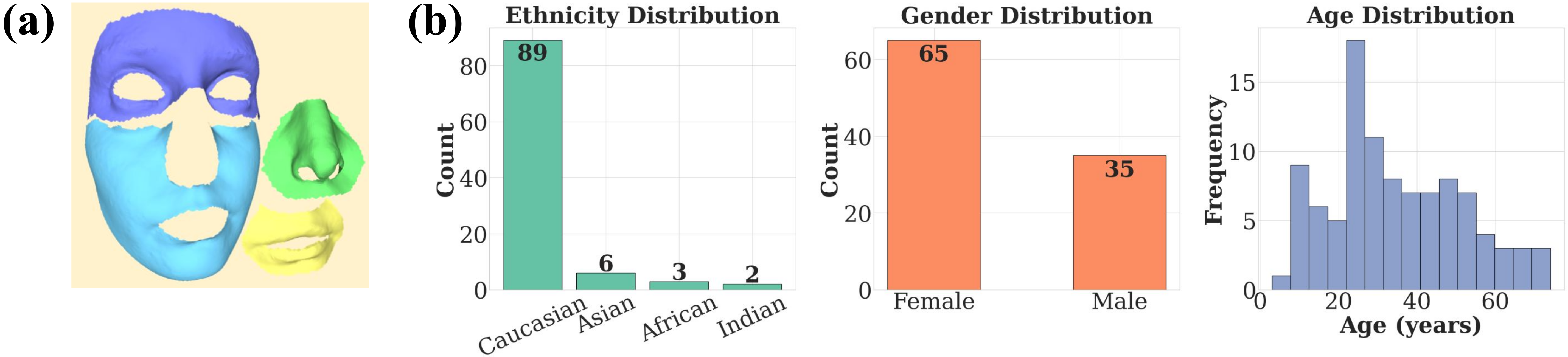}
  \caption{Overview of REALY benchmark: (a) REALY region masks, (b) REALY Demographic statistics}
  \label{fig:realy-demographics}
\end{figure}

For a high-fidelity evaluation of morphological preservation, we choose REALY benchmark \cite{chai2022realy}, a Region-aware benchmark based on the LYHM \cite{dai2020statistical} dataset. Unlike standard 3D datasets, REALY provides 100 globally aligned 3D face scans with accurate facial keypoints and high-quality region masks (nose, mouth, cheek and forehead). Besides, these faces are assigned with demographic labels: age, gender and ethnicity (see \cref{fig:realy-demographics} for details).


\subsubsection{3DMM Base Representativeness.}

In the following sections, we evaluate the representation capacity of the Basel Face Model (BFM) \cite{paysan20093d} and FLAME \cite{li2017learning}, the two most prevalent 3DMMs underpinning current state-of-the-art methods.

To evaluate this representativeness, given a 3D face scan $\mathcal{M}$, we want to find the best face $\mathcal{M}' = (\mathcal{V}', \mathcal{F}')$ that can be expressed in the 3DMM base, written $\mathcal{V}' = \vec{\mu} + \sum_{j=1}^{n} \beta_j \vec{\psi_j}$
with $\vec{\mu}$ the mean face and $\vec{\psi_j}$ the PCs of the 3DMM basis. Note that the reference 3DMM mean shape $\vec{\mu}$ has been preliminarily rigidly aligned and scaled to $\mathcal{M}$ using the ICP.
We obtain the $\beta_j$ values by solving using gradient descent the following optimization problem
\begin{equation}
\label{eq:min_3dmm}
    \min_{\{\beta_j\}_{j=1}^n} \sum_{\vec{v} \in \mathcal{V}} || \vec{v} -  T^{vtx}_{\mathcal{M} \rightarrow \mathcal{M}'}(\vec{v})||^2,
\end{equation}

%


By minimizing this 3D-to-3D distance, we obtain the optimal approximation of the target geometry within the span of the 3DMM basis, allowing us to conclude directly on the model's intrinsic expressive capacity, agnostic of any reconstruction method (e.g. from a 2D picture).

\subsubsection{Reconstruction Algorithms from 2D Image.}

We extended our evaluation to state-of-the-art 3D face reconstruction algorithms from 2D face picture that use 3DMMs as their shape priors: Deep3D\cite{deng2019accurate}, 3DDFA-v3\cite{wang20243d}, MGCNet\cite{shang2020self}, DECA\cite{feng2021learning} and MICA\cite{zielonka2022towards}. To do so, we used as input the frontal face image of each panelist as provided by REALY dataset. Please note that in this case, the ground truth mesh $\mathcal{M}$ is only used to calculate the reconstruction error.

\subsubsection{Anatomical Zones Segmentation.}
In order to perform our evaluation region by region, we can leverage the anatomical zones defined in REALY, which are high-quality masks for the nose, mouth, cheeks, and forehead. However, we need to extend these zones into the 3DMM base, that is to define a list of indices for vertices $v_i$ corresponding to each region. To do so, we propose a \textit{Topological Voting Procedure} to assign regions to the vertices of the 3DMM mean shape $\vec{\mu}$.

To do so, given a 3DMM, we consider its template $\vec{\mu}$ and aim at assigning a region label to each of its vertex $\vec{v}_i$, that can be later used for any face expressed in this base. Given a ground truth scan $\mathcal{M} = (\mathcal{V}, \mathcal{F})$ in the REALY dataset, we first align it with the mesh of $\vec{\mu}$ via the ICP algorithm \cite{besl1992method,amberg2007optimal}. Then, for every vertex $\vec{v}_i$ in the mesh of $\vec{\mu}$, we identify its regional assignment $l(\vec{v}_i,\mathcal{M})$ by finding its closest vertex in $\mathcal{M}$:
\begin{equation}
\label{eq:region_label}
    l(\vec{v}_i,\mathcal{M}) = \text{Label}\left( \arg\min_{\vec{v} \in \mathcal{V}} \| \vec{v}_i - \vec{v} \|^2 \right),
\end{equation}
where $\text{Label}(\cdot)$ returns the anatomical zone (\eg, $\{\text{Nose},\text{Mouth}, \dots\}$) as labeled vertex by vertex in REALY dataset.
We compute similarly this region mask $l(\vec{v}_i,\mathcal{M})$ for every face $\mathcal{M}$ of REALY, and finally assign to $v_i$ the region label that has been the most frequently assigned. It provides a more robust region label to $v_i$ vertex and solves the potential inconsistency of Equation \ref{eq:region_label} from one ground-truth face to another.
This consensus-based approach results in a standardized partition of the 3DMM topology that is anatomically consistent with the REALY definition of morphological zones.

\subsection{Evaluation of the Curvature-Based Metric}
\subsubsection{Perceptual Validation.}


In order to assess whether our proposed \textbf{Curvature Reconstruction Error (CRE)} aligns more closely with human judgment than traditional Euclidean metrics, we conducted a user study. 19 annotators from diverse demographic backgrounds evaluated the reconstruction fidelity in a pairwise comparison.

In detail, for each ground truth scans $\mathcal{M}$ of nose regions from REALY, we reconstructed $\mathcal{M}'$ in Basel Face Model (BFM) using Equation \ref{eq:min_3dmm}.
The users were presented with the following pair at a time, $((\mathcal{M}^A, \mathcal{M}'^A), (\mathcal{M}^B, \mathcal{M}'^B))$, with $A$ and $B$ drawn randomly among a set of $2,414$ pairs. They voted which reconstruction between A and B exhibited higher morphological fidelity relative to its respective ground truth. This pairwise comparative setup allows us to determine whether the reconstruction error comparison provided by a metric $\mathcal{E}$ (i.e., whether $\mathcal{E}(\mathcal{M}^A, \mathcal{M}'^A) < \mathcal{E}(\mathcal{M}^B, \mathcal{M}'^B)$) aligns with human perceptual preference.
For each pair $A$, $B$, we obtained $n_{(A,B)}$ annotations ($n_{(A,B)} \geq$ 5), on which we computed the ratio of agreement. We retained as labeled pairs the $1,111$ pairs for which this ratio exceeds $75\%$.


Given a metric $\mathcal{E}$, we computed its accuracy as the probability that its choice matches the human majority consensus
\begin{equation*} \text{Pairwise-Accuracy}(\mathcal{E}) = \frac{1}{N_{\text{pairs}}}\sum_{j=1}^{N_{\text{pairs}}}\mathds{1}_{\big(\mathcal{E}(\mathcal{M}^A, \mathcal{M}'^A) - \mathcal{E}(\mathcal{M}^B, \mathcal{M}'^B)\big)  P_j\geq 0}\end{equation*}
where $P_j \in \{-1,1\}$ is the perceptual ground truth for $j$-th pair.

\subsubsection{Comparison to Existing Methods and Ablation Study.}

\cref{tab:ablation} summarizes results of our user experiment. We compared CRE with NMSE based on vertex-to-vertex and point-to-surface alignments, and on bi-directional alignment protocol proposed by REALY \cite{chai2022realy} authors. Our curvature-based metric achieves a perceptual accuracy of \textbf{73.63\%} (for $R=3$ and $K=128$), significantly surpassing the performance of NMSE-based benchmarks, which hover near the level of random chance ($\approx 0.5$ for binary choices). This significant margin confirms that local surface character, as captured by CRE, is a much more salient indicator of reconstruction quality than global point-wise distances, making it an ideal candidate for detecting subtle demographic biases in 3DMMs.


To further motivate CRE and justify our design choices, we conduct an extensive ablation study varying $R$ and $K$, and comparing $L_1$, $L_2$, and Huber norms in \cref{eq:cre-metric}. As shown in the lower portion of \cref{tab:ablation}, CRE is robust to its key hyperparameters; remaining configurations with similar trends are reported in the Supplementary Material. \cref{fig:qualitative-comparison} qualitatively contrasts CRE and NMSE on three representative pairs. CRE aligns with human judgments in the first two and diverge on the third one.

\begin{figure}[tb]
  \centering
    \includegraphics[width=0.85\textwidth]{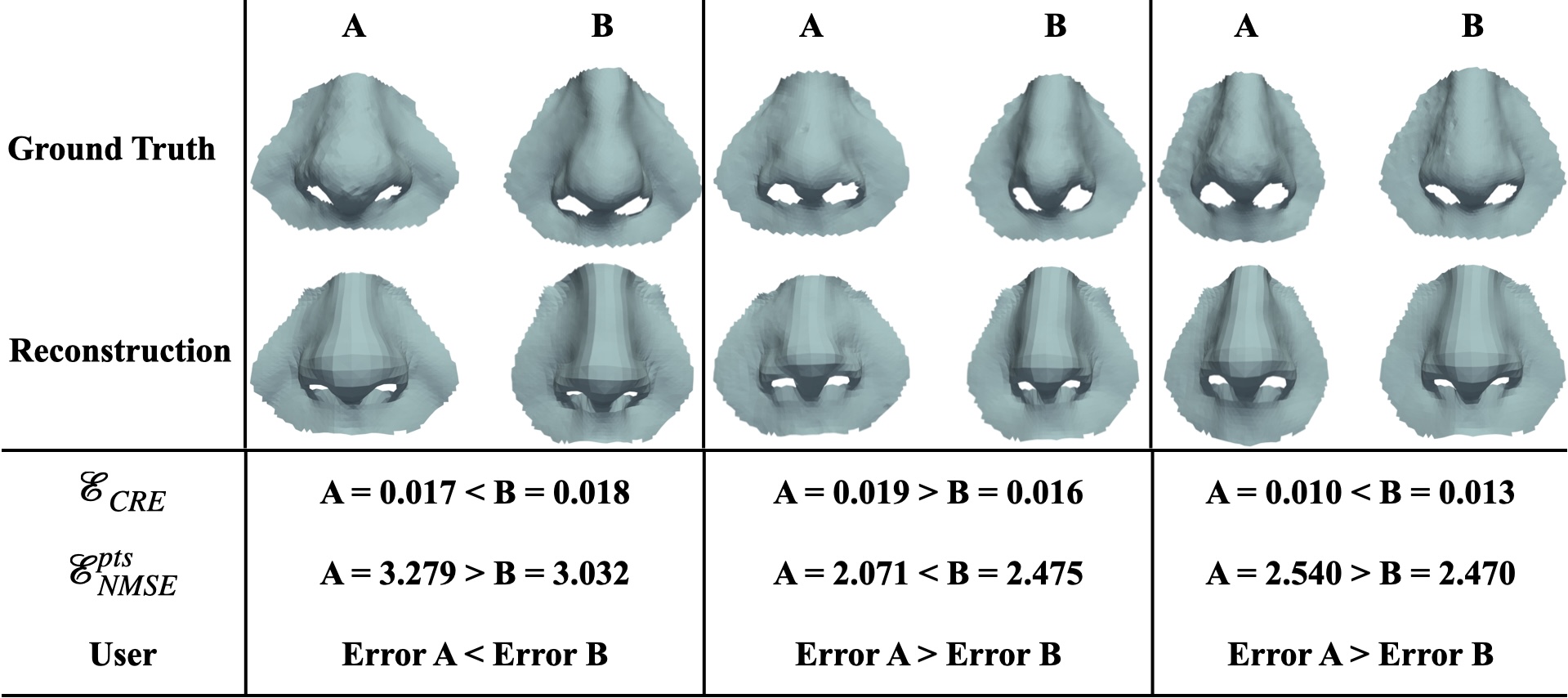}
  \caption{Example of three pairs with their metrics and the user choice.}
  \label{fig:qualitative-comparison}
\end{figure}

\begin{table}[tb]
\caption{Correspondence of reconstruction metrics with human judgments. Top: comparison with existing metrics. Bottom: ablation over $\mathcal{E}_{CRE}$ hyperparameters.
}
\label{tab:ablation}
\centering
    \fontsize{7pt}{10pt}\selectfont
\begin{tabular}{c|c|c|c|c|c|c}
\toprule
Metric       & Mapping       & $R$ & $K$ & Norm  & Pairwise-Accuracy, \% & Pairwise-ROC-AUC, \% \\ \midrule
$\mathcal{E}_{NMSE}^{REALY}$ & REALY & --   & -- & -- & 53.92 &  56.53 \\
\midrule
$\mathcal{E}_{NMSE}^{pts}$ &Points-to-surface & --   & -- & -- & 49.32 &  55.02 \\
\midrule
$\mathcal{E}_{NMSE}^{vtx}$ &Vertex-to-vertex& --   & -- & -- & 42.03 &  42.47 \\
\midrule
$\mathcal{E}_{CRE}$  & Points-to-surface  & \textbf{3} & \textbf{128} & $L_1$ & \textbf{73.63} & \textbf{82.66} \\
\midrule
$\mathcal{E}_{CRE}$ & Points-to-surface & 1   & 128 & $L_1$ &  71.38     &  80.63      \\
$\mathcal{E}_{CRE}$ & Points-to-surface & 1   & 512 & $L_1$ &  68.32    & 78.17        \\
$\mathcal{E}_{CRE}$ & Points-to-surface & 1   & 1024 & $L_1$&  66.52   &  74.35        \\
\midrule
$\mathcal{E}_{CRE}$ & Points-to-surface & 2   & 128 & $L_1$ &  72.46   & 81.86 \\
$\mathcal{E}_{CRE}$ & Points-to-surface & 2   & 512 & $L_1$ &  68.77   & 77.46 \\
$\mathcal{E}_{CRE}$ & Points-to-surface & 2   & 1024 & $L_1$ & 66.79   & 77.46 \\
\midrule
$\mathcal{E}_{CRE}$ & Points-to-surface & 3   & 512 & $L_1$ &  70.66    & 79.36 \\
$\mathcal{E}_{CRE}$ & Points-to-surface & 3   & 1024 & $L_1$ &  69.85   & 78.83 \\
\midrule
$\mathcal{E}_{CRE}$ & Points-to-surface & 4   & 128 & $L_1$ &  73.00    &  81.85 \\
$\mathcal{E}_{CRE}$ & Points-to-surface & 4   & 512 & $L_1$ & 70.93    &  79.40  \\
$\mathcal{E}_{CRE}$ & Points-to-surface & 4   & 1024 & $L_1$ &  70.66   &  79.17 \\
\midrule
$\mathcal{E}_{CRE}$ & Points-to-surface & 3   & 128 & $L_2$ &   72.28   &  81.17 \\
$\mathcal{E}_{CRE}$ & Points-to-surface & 3   & 512 & $L_2$ &  69.67   & 77.81  \\
$\mathcal{E}_{CRE}$ & Points-to-surface & 3   & 1024 & $L_2$ & 68.14   &  76.99 \\
\midrule
$\mathcal{E}_{CRE}$ & Points-to-surface & 3   & 128 & \textit{Huber} &  67.60      &     75.82   \\
$\mathcal{E}_{CRE}$ & Points-to-surface & 3   & 512 & \textit{Huber} &  62.74     &    70.40      \\
$\mathcal{E}_{CRE}$ & Points-to-surface & 3   & 1024 & \textit{Huber} & 61.57   &     69.51    \\
\bottomrule
\end{tabular}
\end{table}

\subsection{Biases and Fairness Analysis}\label{ssec:FairnessEvaluation}

In this section, we use the proposed Curvature Reconstruction Error and regional spectral analysis to quantify demographic biases in 3DMM bases. By analyzing the relationship between reconstruction fidelity and sensitive attributes, we uncover systematic disparities in how these models represent diverse populations.

\subsubsection{Age Bias with respect to CRE.}

We first investigate the correlation between subject age and the fidelity of the 3DMM representation. We hypothesize that as facial morphology becomes more complex with age (\eg, due to the appearance of folds, creases, and sagging), the linear basis of standard 3DMMs becomes increasingly insufficient. To quantify this, we compute Pearson $r$ correlation coefficient \cite{myers2013research} between subject age and the regional $\mathcal{E}_{CRE}$.

\begin{table}[tb]
\caption{Pearson correlation coefficients between age and CRE across different regions. Strong correlations ($\geq 0.5$) are highlighted in bold.}
  \label{tab:correlation-3dmm}
\centering
    \fontsize{7pt}{10pt}\selectfont
\begin{tabular}{ll|cccccccccccc}
    \toprule
      & & \multicolumn{12}{c}{Pearson correlation: $\mathcal{E}_{CRE}$ $\vs$ 3D Reconstruction}                                                                                                                                                                                                                                                                                                                     \\
      & & \multicolumn{3}{c|}{@nose}                                                                                   & \multicolumn{3}{c|}{@mouth}                                                                   & \multicolumn{3}{c|}{@cheek}                                                                                  & \multicolumn{3}{c}{@forehead}                                                    \\ 
      & & \multicolumn{1}{c|}{All}           & \multicolumn{1}{c|}{Female}        & \multicolumn{1}{c|}{Male}          & \multicolumn{1}{c|}{All}   & \multicolumn{1}{c|}{Female} & \multicolumn{1}{c|}{Male}          & \multicolumn{1}{c|}{All}           & \multicolumn{1}{c|}{Female}        & \multicolumn{1}{c|}{Male}          & \multicolumn{1}{c|}{All}           & \multicolumn{1}{c|}{Female} & Male          \\ \midrule
\textit{3DMM} & BFM   & \multicolumn{1}{c|}{0.39}          & \multicolumn{1}{c|}{0.32}          & \multicolumn{1}{c|}{\textbf{0.59}} & \multicolumn{1}{c|}{0.33}  & \multicolumn{1}{c|}{0.21}   & \multicolumn{1}{c|}{\textbf{0.67}} & \multicolumn{1}{c|}{\textbf{0.77}} & \multicolumn{1}{c|}{\textbf{0.72}} & \multicolumn{1}{c|}{\textbf{0.88}} & \multicolumn{1}{c|}{\textbf{0.52}} & \multicolumn{1}{c|}{0.49}   & \textbf{0.59} \\
\textit{Repr.} & FLAME & \multicolumn{1}{c|}{\textbf{0.78}} & \multicolumn{1}{c|}{\textbf{0.79}} & \multicolumn{1}{c|}{\textbf{0.78}} & \multicolumn{1}{c|}{-0.03} & \multicolumn{1}{c|}{-0.16}  & \multicolumn{1}{c|}{0.18}          & \multicolumn{1}{c|}{\textbf{0.82}} & \multicolumn{1}{c|}{\textbf{0.84}} & \multicolumn{1}{c|}{\textbf{0.82}} & \multicolumn{1}{c|}{0.41}          & \multicolumn{1}{c|}{0.48}   & 0.25     \\
\midrule
& Deep3D   & \multicolumn{1}{c|}{0.24}      & \multicolumn{1}{c|}{0.18}      & \multicolumn{1}{c|}{0.41} & \multicolumn{1}{c|}{-0.23} & \multicolumn{1}{c|}{-0.19}  & \multicolumn{1}{c|}{-0.38} & \multicolumn{1}{c|}{\textbf{0.69}} & \multicolumn{1}{c|}{\textbf{0.81}} & \multicolumn{1}{c|}{\textbf{0.62}} & \multicolumn{1}{c|}{\textbf{0.58}} & \multicolumn{1}{c|}{\textbf{0.54}}   & \textbf{0.66} \\
\textit{Model} & 3DDFA-v3 & \multicolumn{1}{c|}{\textbf{0.67}} & \multicolumn{1}{c|}{\textbf{0.65}} & \multicolumn{1}{c|}{\textbf{0.73}}     & \multicolumn{1}{c|}{0.31}      & \multicolumn{1}{c|}{0.30}       & \multicolumn{1}{c|}{0.37}               & \multicolumn{1}{c|}{\textbf{0.79}}     & \multicolumn{1}{c|}{\textbf{0.78}}     & \multicolumn{1}{c|}{\textbf{0.88}}     & \multicolumn{1}{c|}{\textbf{0.53}}              & \multicolumn{1}{c|}{\textbf{0.55}}       &   \multicolumn{1}{c}{0.48}            \\
\textit{from} & MGCNet  & \multicolumn{1}{c|}{\textbf{0.54}}          & \multicolumn{1}{c|}{\textbf{0.50}}          & \multicolumn{1}{c|}{\textbf{0.67}}              & \multicolumn{1}{c|}{0.30}      & \multicolumn{1}{c|}{0.30}       & \multicolumn{1}{c|}{0.32}               & \multicolumn{1}{c|}{\textbf{0.80}}              & \multicolumn{1}{c|}{\textbf{0.79}}              & \multicolumn{1}{c|}{\textbf{0.89}}              & \multicolumn{1}{c|}{\textbf{0.51}}              & \multicolumn{1}{c|}{\textbf{0.54}}       &      \multicolumn{1}{c}{0.47}         \\
\textit{2D image} & DECA   & \multicolumn{1}{c|}{\textbf{0.73}}          & \multicolumn{1}{c|}{\textbf{0.75}}          & \multicolumn{1}{c|}{\textbf{0.70}}              & \multicolumn{1}{c|}{-0.32}      & \multicolumn{1}{c|}{-0.30}       & \multicolumn{1}{c|}{-0.35}               & \multicolumn{1}{c|}{\textbf{0.76}}              & \multicolumn{1}{c|}{\textbf{0.80}}              & \multicolumn{1}{c|}{\textbf{0.85}}              & \multicolumn{1}{c|}{0.40}              & \multicolumn{1}{c|}{0.46}       &       \multicolumn{1}{c}{0.28}        \\
& MICA   & \multicolumn{1}{c|}{\textbf{0.69}}          & \multicolumn{1}{c|}{\textbf{0.72}}          & \multicolumn{1}{c|}{\textbf{0.65}}              & \multicolumn{1}{c|}{-0.31}      & \multicolumn{1}{c|}{-0.31}       & \multicolumn{1}{c|}{-0.30}               & \multicolumn{1}{c|}{\textbf{0.79}}              & \multicolumn{1}{c|}{\textbf{0.76}}              & \multicolumn{1}{c|}{\textbf{0.88}}              & \multicolumn{1}{c|}{0.43}              & \multicolumn{1}{c|}{0.48}       &   \multicolumn{1}{c}{0.31}  \\
\bottomrule
\end{tabular}
\end{table}

\cref{tab:correlation-3dmm} summarizes these findings for the BFM and FLAME models, and the 3D reconstruction models from images. For 3DMMs, we observe a strong positive correlation between age and reconstruction error across nearly all anatomical regions. Notably, in the cheek region, an area highly susceptible to age, related morphological changes -- the Pearson correlation reaches \textbf{0.77} for BFM and \textbf{0.82} for FLAME.
We also see for 3D reconstruction models that the age-related bias is consistently propagated from the 3D prior to the final output. Notably, however, this bias do not seem exacerbated by these predictive algorithms.

\begin{figure}[tb]
    \centering
    \includegraphics[width=\linewidth]{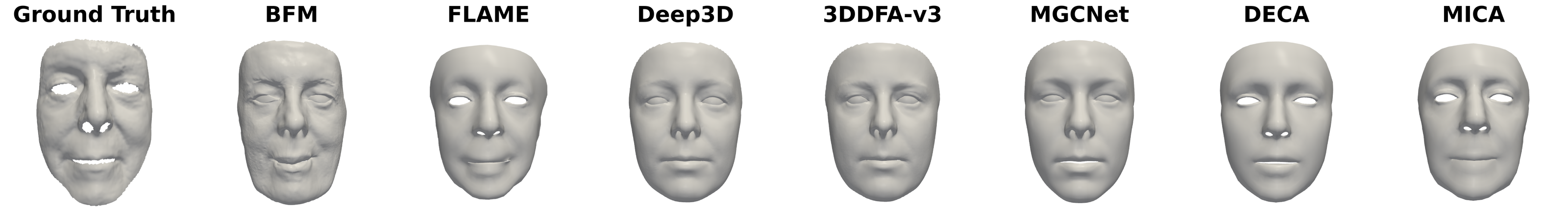}
    \caption{Example of age bias on a senior person from REALY benchmark.}
    \label{fig:age-bias}
\end{figure}

These results reveal a significant representation bias: as age increases, the 3DMM basis and reconstruction models fails to capture high-frequency geometric details, effectively "smoothing out" the structural markers characteristic of older individuals. \cref{fig:age-bias} illustrates such effects. Moreover, gender-stratified analysis shows that this age-related degradation is prevalent in both populations, though it is notably intensified in males within the BFM basis (\eg, $r=0.88$ for the cheek region).
To further confirm the benefits of a curvature-based approach, we repeated this analysis using the standard Euclidean-based NMSE and observed no significant bias (see the Supplementary Material).

\subsubsection{Ethnicity with respect to Error Clusters.}

\begin{figure}[tb]
    \centering
    \includegraphics[width=\linewidth]{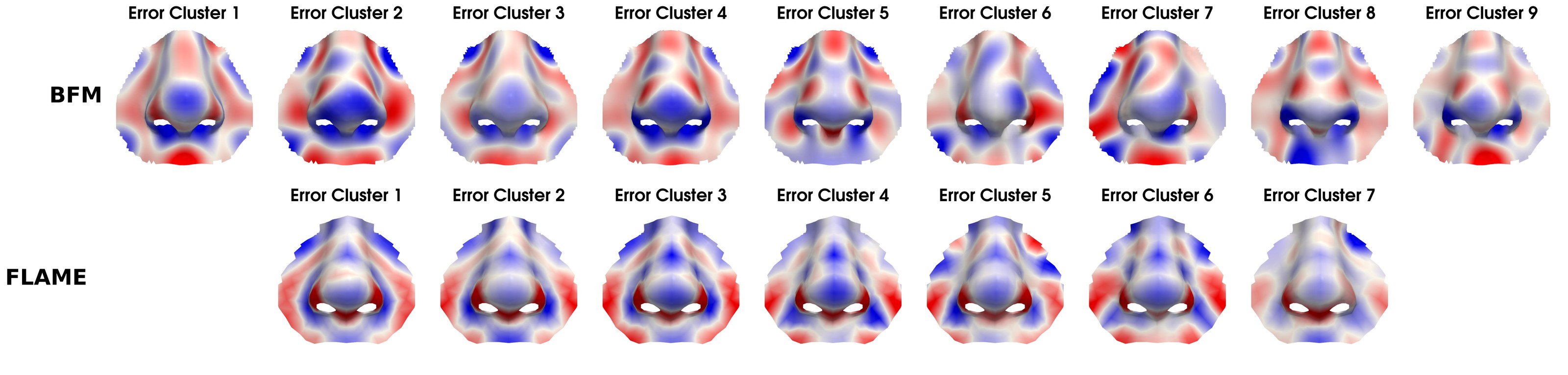}
    \caption{Error cluster centroids found for BFM (top row) and FLAME (bottom row).}
    \label{fig:nose-error-cluster-centroids}
\end{figure}

\begin{figure}[tb]
  \centering
  \begin{subfigure}{0.48\linewidth}
    \includegraphics[width=0.8\textwidth]{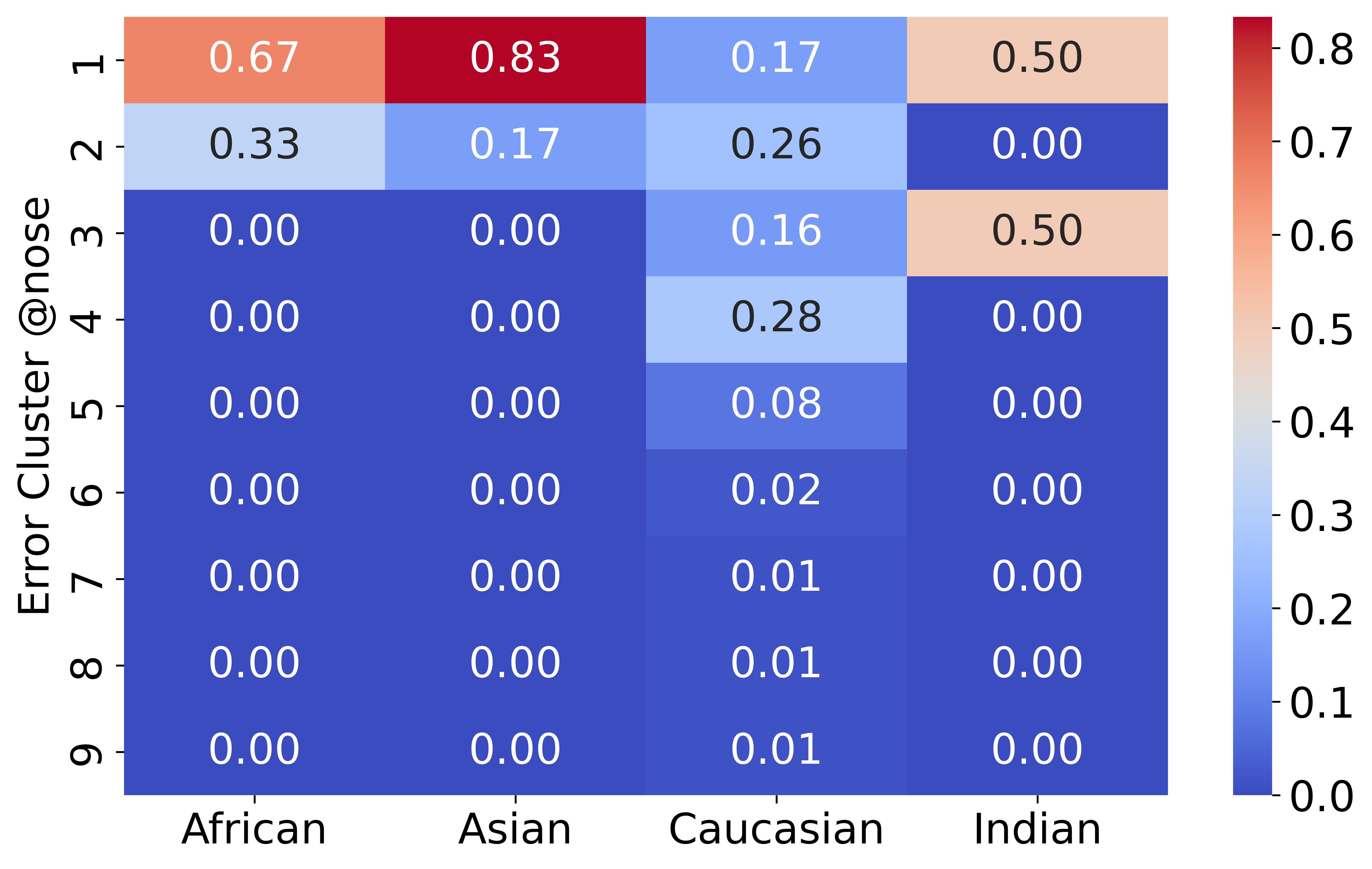}
    \caption{BFM \cite{paysan20093d}}
    \label{fig:bfm-error-clusters-nose}
  \end{subfigure}
  \hfill
  \begin{subfigure}{0.48\linewidth}
    \includegraphics[width=0.8\textwidth]{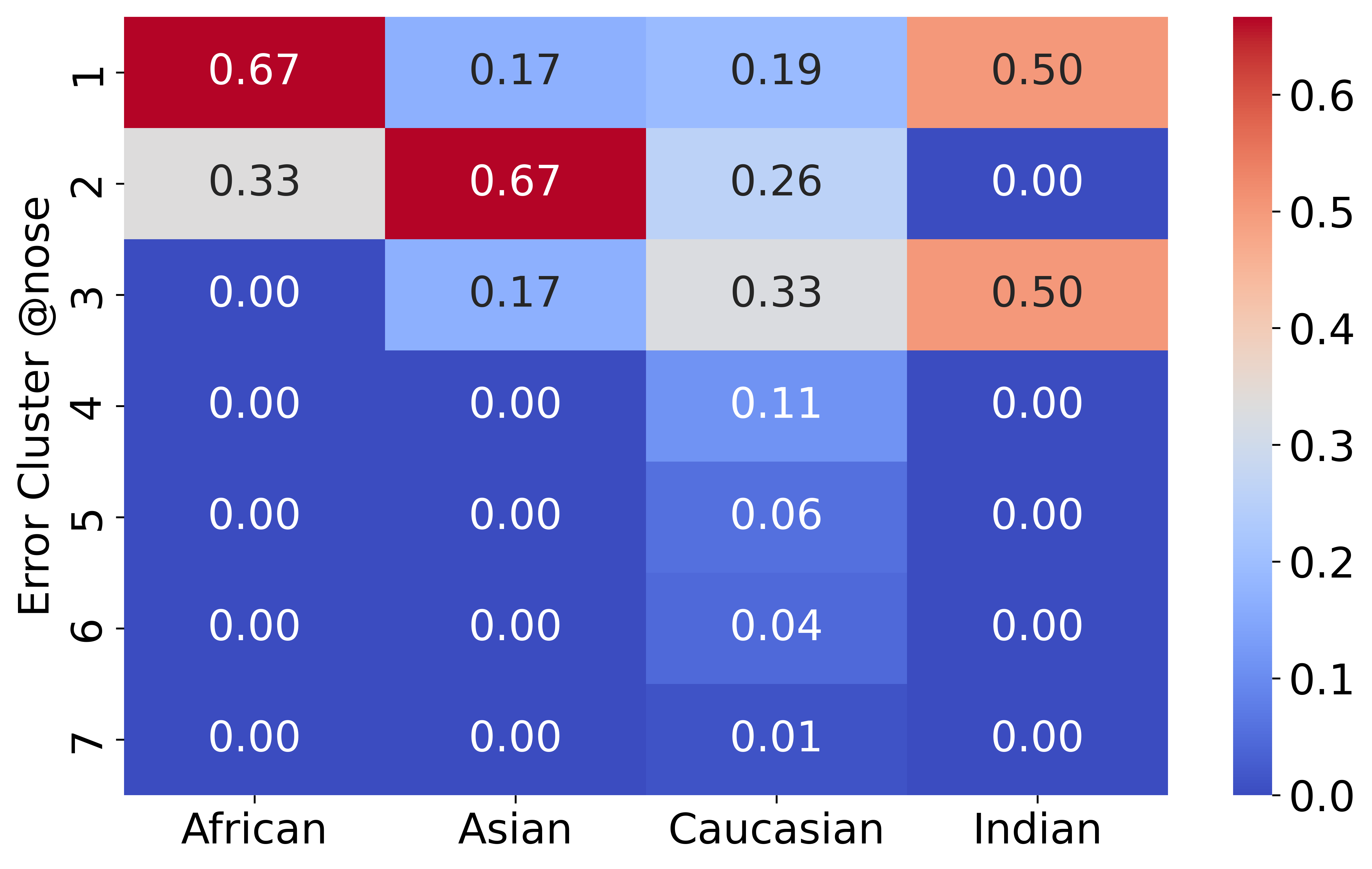}
    \caption{FLAME \cite{li2017learning}}
    \label{fig:flame-error-clusters-nose}
  \end{subfigure}
  \caption{Distribution of failure modes @nose for each ethnicity in REALY benchmark.}
  \label{fig:error-clusters-nose}
\end{figure}

\begin{figure}[tb]
  \centering
    \includegraphics[width=\textwidth]{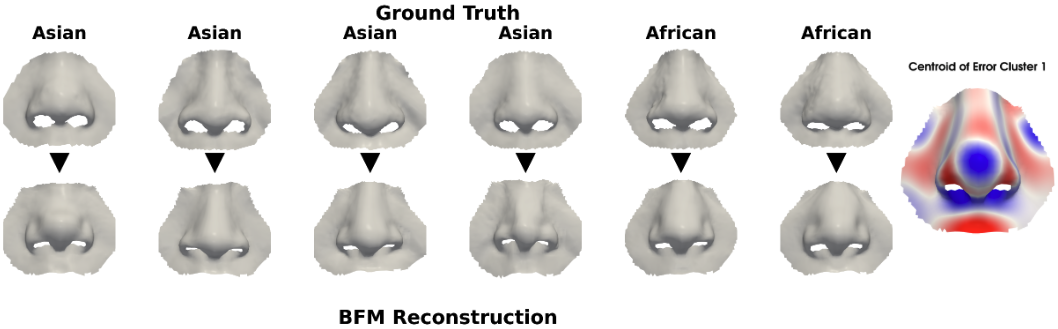}
   \caption{Example of African and Asian subjects affected by a failure mode 1 of BFM \cite{paysan20093d}.}
  \label{fig:error-pattern-nose}
\end{figure}

Given the very limited number of samples per ethnicity (except Caucasians), we moved beyond global error analysis and clustered regional error maps $\epsilon_H$ as described in \cref{alg:kmeans-algo}. 

\cref{fig:nose-error-cluster-centroids} illustrates 9 (resp. 7) unique failure modes that were identified for BFM (resp. FLAME) reconstructions for the nose zone (see Supplementary Material for other anatomical zones). \cref{fig:error-clusters-nose} summarizes the distributions of these failure modes for each ethnicity. Whereas Caucasian subjects are distributed across different error clusters, African and Asian populations are primarily concentrated at the first error cluster for BFM and at the first and second error clusters for FLAME. This might suggest that these populations suffer from a distinct systematic error pattern produced by these 3DMMs. \cref{fig:error-pattern-nose} illustrates this pattern for BFM reconstructions. We observe that the nose wings of reconstructions have unnatural creases and the tip of the nose is drawn up and more round.
%
The linear subspaces of both these 3DMMs lack the requisite expressive capacity to capture the distinct geometric nuances found in globally diverse populations, leading to a systematic "flattening" of ethnic-specific features toward the model mean.

\subsubsection{Morphology Traits with respect to Error Clusters.}

\begin{figure}[htb]
  \centering
  \includegraphics[width=1\linewidth]{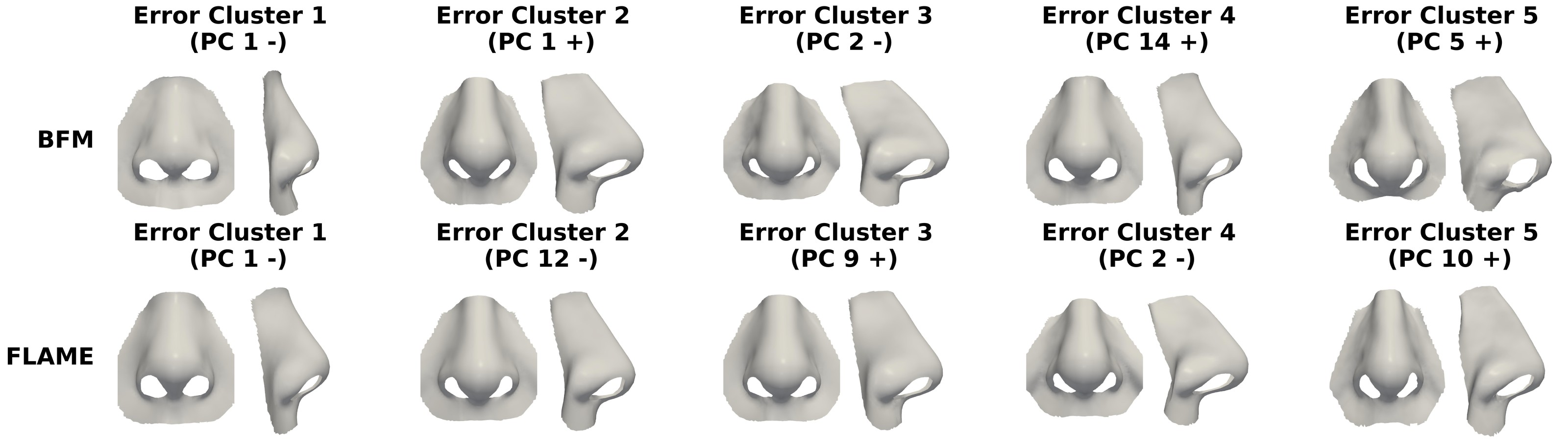}
  \caption{Centroids for error clusters identified for BFM \cite{paysan20093d} and FLAME \cite{li2017learning}. In $({\cdot})$ we put the $PC$ with maximal mean absolute value for the error cluster and its sign.}
  \label{fig:pc-centroids-morphology-vs-error}
\end{figure}

To demonstrate that 3DMM failure modes are systematic rather than stochastic, we investigate the intrinsic relationship between specific morphologies and curvature-based reconstruction errors.
To visualize the morphologies associated to each error cluster, we computed a PCA on the REALY ground truth faces. Such decomposition mirrors the statistical formulation used to construct a 3DMM basis, with the difference that we focused on a region of interest (more details are given in the Supplementary Material).
For each error cluster, we identify the dominant principal component -- the dimension that primarily governs the cluster's morphology -- defined as the component with the maximal mean absolute coefficient value. Furthermore, we analyze the sign of these mean coefficients to determine the direction of morphological deviation relative to the mean shape. 

The results for five representative error clusters in both BFM and FLAME are illustrated in \cref{fig:pc-centroids-morphology-vs-error} (refer to the Supplementary Material for an exhaustive analysis). Our observations confirm that each error pattern maps to a distinct, well-defined morphological phenotype. For example, within the BFM framework, Error Cluster 1 is driven by $PC_1$, which encodes variations in nasal length. A negative coefficient sign indicates that this cluster is predominantly composed of individuals with significantly shorter nasal structures than the population mean. This direct mapping suggests that 3DMM biases are not random artifacts of the fitting process, but are fundamentally rooted in the model's inability to span specific regions of the morphological manifold.

\section{Discussion and Conclusion}

We proposed a new curvature-aware paradigm for analyzing 3D face reconstruction errors, enabling us to define: (1) an error map for localized geometric visualization; (2) a scalar metric that is significantly better aligned with human perception than standard Euclidean-based metrics; and (3) a clustering of errors to discover unknown biases.

Our fairness analysis reveals that both evaluated 3DMMs exhibit pronounced age-related biases, even in the idealized case of projecting the full 3D geometric ground truth into the 3DMM basis. We further find preliminary evidence of gender, ethnic, and morphological biases. This suggests that the reconstruction models' representational bottlenecks are inherent to their linear prediction subspaces rather than a failure of 2D-to-3D inference, perpetuating demographic disparities regardless of the fitting methodology.

As future work, we plan to investigate how our error map can be used as a differentiable loss function for a neural network for 3D reconstruction. This curvature-aware objective would drive a secondary refinement module, in order to go beyond the raw 3DMM basis limitations.



%
%
\bibliographystyle{splncs04}
\bibliography{main}
\end{document}